\documentclass[11pt]{article}
\usepackage{coling2020}
\usepackage{times}
\usepackage{url}
\usepackage{latexsym}
\usepackage{graphicx}
\usepackage{comment}
\usepackage{booktabs}
\usepackage{amsmath}
\usepackage{multirow}
\usepackage{soul, xcolor}

\colingfinalcopy % Uncomment this line for the final submission

\title{Downstream Transformer Generation of Question-Answer Pairs with Preprocessing and Postprocessing Pipelines}
\author{Cheng Zhang \and Hao Zhang \and Jie Wang \\
  Department. of Computer Science \\
  University of Massachusetts,
  Lowell, MA 01854\\
  \texttt{\{cheng\_zhang, hao\_zhang\}@student.uml.edu, wang@cs.uml.edu}}

\date{}

\begin{document}
\maketitle

\begin{abstract}
We present a system called TP3 to perform a downstream task of transformers on generating question-answer pairs (QAPs) from a given article. TP3 first finetunes pretrained transformers on QAP datasets, then uses a preprocessing pipeline to select appropriate answers, feeds the relevant sentences and the answer to the finetuned transformer to generate candidate QAPs, and finally uses a postprocessing pipeline to filter inadequate QAPs. In particular,
using pretrained T5 models as transformers and the SQuAD dataset as the finetruning dataset, we show that
TP3 generates satisfactory number of QAPs with high qualities on the Gaokao-EN dataset. 
\end{abstract}

\section{Introduction}

%Question generation is a challenge task because it requires first a good understanding of the context, then requires how to establish a relation between the context and answer, and finally requires the model the ability to generate the questions fluently as a human.

It is a grand challenge to generate adequate question-answer pairs (QAPs) from a given article. % in NLP.
A QAP is adequate if both the question and the answer are contextually and grammatically correct and %are both syntactically and semantically correct and 
conform to native speakers, while the answer matches the question in the context of the article.
%which aims to generate natural and relevant questions based on given contents, 
%where the generated questions need to be able to be answered by the contents. 
%It 
%Such tasks have a variety of applications, ranging from improving the performance of a question-answer system \cite{duan-etal-2017-question} to generating queries for a search engine \cite{zhao-etal-2011-automatically}, 
%and from guiding a chatbot to lead a conversation \cite{wang-etal-2018-learning-ask} to enhancing reading comprehension drills \cite{metaseq-2020}, to name just a few.
%
%We study automatic generation of question-answer pairs (QAPs) with an emphasis on grammatical correctness of the questions and the suitability of the answers. By grammatical correctness we mean that the questions being generated are syntactically and semantically correct and conform to what a native speaker would say.
%
Existing methods on question generation are based either on handcrafted features 
or on deep neural-network models. Methods of the former typically rely on grammar rules. However, no matter how many rules are formatted and used, there are always exceptions that these rules don't apply, resulting in generations of inadequate questions. Methods of the latter typically perform better, % the neural networks based methods 
%generating well-formed and grammatically correct questions from the given answer and context, it 
but may still generate incoherent questions. For instance, asking about %the type of verbs or 
clauses that express reasons or purposes may produce incoherent QAPs.
%While meeting with certain success from different perspectives, the grand challenge of generating adequate QAPs still remains.

We present TP3 (Transformer with Preprocessing and Postprocessing Pipelines) for %, a system %explore and experiment the utilization of 
generating QAPs. It is a downstream task on a pretrained transformer that is 
finetuned on a QAP dataset, with
a preprocessing pipeline to select appropriate answers and a postprocessing pipeline to filter undesirable %and irrelevant 
questions.
These pipelines are a combination of various NLP tools and algorithms.
%T5 is a transformer that uses a text-to-text approach. We use the finetuned SQuAD dataset on pretrained T5 model for our downstream task of question generation. 
%As T5 model performs well on standard NLP tasks, we proposed a system where we used T5 model to generate adequate QAPs candidate questions, and we then use a combination of various NLP tools and algorithms, designed a preprocessing pipeline to select appropriate answers and a postprocessing pipeline to filter the undesirable and irrelevant questions.
In particular, we finetune a pretrained T5-Large model (a Text-to-Text Transfer Transformer pretrained on a large dataset) \cite{raffel2020exploring} on the SQuAD dataset \cite{Rajpurkar_2016} and show that it outperforms the state-of-the-art results under standard metrics. 

For simplicity, we still use TP3 to denote this finetuned T5 transformer with
the preprocessing and postprocessing pipelines. We show that
%TP3-SQuAD, based on a pretrained T5 that is finetuned over SQuAD \cite{}, outperforms common models both under standard metrics and human judgments.
%In particular, TP3-Large-SQuAD (finetuned T5-Large over SQuAD) 
%TP3, with the learning rate of 6e-6 for T5-Large, 
it generates
over 92\% adequate QAPs among the 1,172 QAPs generated on the Gaokao-EN dataset of 85 articles of length
ranging from 15 to 20 sentences each, collected from multiple Gaokao English tests for college entrance examinations. 
% with respect to common evaluation metrics on generating
%Shown in the performance evaluation in section \ref{sec:evaluations}, our QG system 
%is able to generate 
%adequate QAPs %well-formed and grammatically correct questions 
%from the input text. %, outperforms other compared models in most evaluation metrics.

%Our goal is that the QG system can generate questions with a correct rate of more than 90\%.

The rest of the paper is organized as follows:
We discuss in Section \ref{sec:relatedworks} related work on question generation. 
In Section \ref{sec:finetuning}, we describe how we finetune pretrained T5 models and in Section \ref{sec:P3} we 
present our preprocessing and postprocessing pipelines. We provide in
Section \ref{sec:evaluations} evaluation results and conclude the paper in
Section \ref{sec:conclusions}. %, conclusions and the future scope of the paper is also discussed.

\section{Related Works}\label{sec:relatedworks}

Early methods on question generation are typically rule-based systems
%follow the rule-based transformative approach 
that transform %key phrases from 
a declarative sentence into a factual interrogative sentence \cite{heilman2009question,lindberg2013automatic,labutov-etal-2015-deep}.
%These approaches follow the same basic strategy, include 
This approach includes methods to identify key phrases from input sentences, generate questions and answers using syntactic or semantic parsers and named entity analyzers, and transform declarative sentences into interrogative sentences based on linguistic features and syntactic rules for different types of questions \cite{danon2017syntactic,khullar2018automatic,ali2010automation}.
%These methods are confined by %rely on the knowledge of 
%the authors who design the rules of generating questions.
% and the performance is limited by the expert's knowledge. 

The quality of generated QAPs are evaluated either by human judges or by the following metrics: 
BLEU \cite{10.3115/1073083.1073135}, ROUGE \cite{lin-2004-rouge}, and METEOR \cite{banerjee-lavie-2005-meteor},  
even though
none of these metrics measure grammatical correctness of the questions being generated.
%These metrics, however, do not
% are grammatically correct because they are based on N-grams and do not measure grammatical correctness.

Recent advances of neural-network research provide new tools to build generative models.
For example, the attention mechanism can help determine what content in a sentence should be asked \cite{luong-etal-2015-effective}, and the sequence-to-sequence  \cite{bahdanau2014neural,cho-etal-2014-learning} and the long short-term memory  \cite{Sak2014LongSM}  mechanisms are used to generate words to form a question (see, e.g., \cite{du-etal-2017-learning,duan-etal-2017-question,Harrison_2018,sachan-xing-2018-self}).
These models generate questions without the corresponding correct answers. 
%Moreover, training these models require a dataset comprising over 100K questions.
To address this issue, % the problem of generating questions without answers, 
researchers have explored ways to encode a passage (a sentence or multiple sentences) and an answer word (or a phrase) as input, and determine what questions are to be generated for a given answer \cite{10.1007/978-3-319-73618-1_56,zhao-etal-2018-paragraph,song-etal-2018-leveraging}. 
However, as pointed out by
Kim et al. \cite{Kim_2019}, these methods could generate answer-revealing questions, namely, questions contain in them the corresponding answers. They then devised a new method by encoding answers separately, at the expense of learning substantially more parameters. 

More recently, researchers have explored how to use pretrained transformers to generate answer-aware questions \cite{dong2019unified,Zhang2019AddressingSD,Zhou2019QuestiontypeDQ,qi-etal-2020-prophetnet,su-etal-2020-multi,10.1145/3442381.3449892}.
For example, Kettip et al. \cite{Kriangchaivech190905017} presented an architecture for a transformer to generate questions. Rather than fully encoding the context and answers as they appear in the dataset, they applied certain transformations such as the change of named entities both on the context and the answer. 
Lopez et al. \cite{Lopez2020TransformerbasedEQ} finetuned the pretrained GPT-2 \cite{radford2019language} transformer without using any additional complex components or features to enhance its performance.
Chen \cite{Chen2020ReinforcementLB} described a fully transformer-based reinforcement learning generator evaluator architecture to generate questions.

The recent introduction of T5 \cite{raffel2020exploring} has escalated NLP research in a number of ways.
%, which achieves the state-of-the-art results on several NLP task trained on a newer and larger text corpus. % called C4
%\subsection{An Overview of T5}
%
%We present a deep-learning-based end-to-end question generation model to utilize the power of Text-to-Text Transfer Transformer (T5) \cite{raffel2020exploring}. 
T5 is a encoder-decoder text-to-text transformer using the teacher forcing method on a wide variety of NLP tasks, including text classification, question answering, machine translation, and abstractive summarization. Unlike other transformer models (e.g. GPT-2 \cite{radford2019language}) that take in text data after converting them to corresponding numerical embeddings, T5 handles each task by taking in data in the form of text and producing text outputs. 
% Fig \ref{fig:1} depicts the architecture of T5 . 
%The model has an encoder-decoder based transformer architecture \cite{raffel2020exploring}. 

% \begin{figure}[h]
%   \centering
%   \includegraphics[width= \linewidth]{T5Architecture}
%   \caption{T5 Architecture}
%   \label{fig:1}
% %   \Description{}
% \end{figure}

% In particular,
% the encoder consists of a stack of identical layers. Every layer is composed of two sub-layers. The first sub-layer of each encoder layer is a multi-head self-attention mechanism. The second sub-layer  is a fully connected feed-forward network. Residual connections are employed around these sub-layers, each followed by a normalization layer. 
% %
% The decoder also consists of a stack of identical layers. In addition to the two sub-layers already present in the encoder layer, the third sub-layer performs multi-head attention on the output received by the encoder stack. Residual connections are employed around these sub-layers, such as those in the encoder, each followed by a normalization layer. 

%
%
%most recent innovation on question generation train a %pretrained %the Text-to-Text Transfer Transformer (
%T5 model \cite{raffel2020exploring} %, it pretrained upon 
%on a newer and larger text corpus called C4, which achieves the state-of-the-art results on several NLP tasks.
Taking the advantage of pretrained T5, 
Lidiya et al. \cite{mixqg.2110.08175} combined nine question-answering datasets to finetune a single T5 model and evaluated generated questions using a new semantic measure called BERTScore \cite{bert-score}.
Their method achieves so far the best results. % state-of-art result.
We present a finetuned T5 model on a
single SQuAD dataset %with preprocessing and postprocessing pipelines 
to produce better results. %than the Lidiya's approach.

\section{Description of TP3}\label{sec:finetuning}

We describe how we train and finetune a pretrained T5 transformer for our downstream task of question generation and use a combination of various NLP tools and algorithms to build the preprocessing and postprocessing pipelines for generating QAPs. %the QG system.

%More details can be referred from the original paper \cite{raffel2020exploring}.

%\subsection{Selecting a dataset to finetune T5}

%To use the T5 model for our downstream task of question generation, we finetune the T5 model on QAP datasets.

There are a number of public QAP datasets available for fine-tuning T5, including RACE \cite{lai2017large}, CoQA \cite{reddy2019coqa}, and SQuAD \cite{Rajpurkar_2016}. 
RACE is a large-scale dataset collected from Gaokao English examinations over the years, where Gaokao is the national college entrance examinations held once every year in mainland China. It consists of
 more than 28,000 passages and nearly 100,000 questions, including cloze questions.
CoQA is a conversational-style question-answer dataset. It contains a series of interconnected questions and answers %that appear 
in conversations. % about a text. 
SQuAD is a reading comprehension dataset, consisting of more than 100,000 QAPs %and corresponding answers 
posted by crowdworkers on a set of Wikipedia articles.

Among these datasets, %Based on question types, and the 
SQuAD is more commonly used in the question generation research. We use SQuAD to finetune pretrained T5 models. 
%
%The T5 model is tasked with generating a relevant question for a given answer and a context for that answer.
For each QAP and the corresponding context extracted from the SQuAD training dataset, we concatenate the answer and the context with markings in the format of $\langle answer\rangle answer\_text \langle context\rangle context\_text$ as input, with the question as the target, where the context is the entire article for the QAP in SQuAD.
%for calculating loss during training, 
We then set the maximum input length to 512 and the target length to 128 to avoid infinite loops and repetitions of target outputs. We feed the concatenated text input and question target into a pretrained T5 model for fine-tuning and use AdamW \cite{loshchilov2018decoupled} as an optimizer with various learning rates to obtain a better model.

%We discuss the experiment results in Section \ref{sec:evaluations}
%large auto lr: 0.0019054607179632484
%base auto lr: 0.0004365158322401656
%We'd like to explore a large number of different learning rates. 
To explore various learning rates, we first use automatic evaluation methods to narrow down a smaller range of the learning rates and then use human judges to determine the best learning rate.
In particular, we first finetune the base model with a learning rate of $1.905 \times 10^{-3}$ and the large model with a learning rate of $4.365 \times 10^{-4}$. The learning rates are calculated %selected 
using the Cyclical Learning Rates (CLR) method \cite{7926641}, which is used to find automatically the best global learning rate. % and schedule for the global learning rates.
%, and it is used in many well-known deep learning frameworks, such as PyTorch and TensorFlow.
Evaluated by human judges, we found that %the generated  are fair but not ideal as our expected. We also noticed that 
the best learning rate calculated by CLR is always larger than the actual best learning rate in our experiments.

We then finetune T5-Base and T5-Large with dynamic learning rates from the learning rate calculated by CLR with a reduced learning rate for each epoch. For example, we finetune T5-Base starting from a learning rate of $1.905 \times 10^{-3}$ and multiply the previous learning rate by 0.5 for the current epoch until % multiplication in each epoch decreased to 
the learning rate of $1.86 \times 10^{-6}$ is reached. Likewise, we finetune T5-Large in the same way starting from $4.365 \times 10^{-4}$ until
%,  with 0.5 multiplication in each epoch decreased to 
the learning rate of $1.364 \times 10^{-5}$ is reached. However, the generated results are still below expectations. 

We therefore proceed to 
%Since the learning rate calculated by CLR is not reliable, we then tried to 
finetune the models with various learning rates we choose. In particular,
%According to previous models training experience, 
we first finetune T5-Base with a learning rate from $10^{-3}$ to $10^{-4}$ with a $2.5 \times 10^{-4}$ decrement for each epoch, and from $10^{-4}$ to $10^{-5}$ with a $2.5 \times 10^{-5}$ decrement for each epoch. Likewise, we  
finetune T5-Large with a learning rate from $10^{-4}$ to $10^{-5}$ with a $2 \times 10^{-5}$ decrement for each epoch, and from $10^{-5}$ to $10^{-6}$ with a $2 \times 10^{-6}$ decrement for each epoch.

%The following metrics are used to automatically evaluate the performance of question generation from the T5 models:
%
%BLEU scores measure the quality of text that has been translated by a machine from one natural language to another using n-grams. We used a cumulative 4-gram BLEU score (B4) as an evaluation metric.
%
%ROUGE-L uses statistics based on the Longest Common Subsequence (LCS) to evaluate recall by how many words in reference sentences are used in predicted sentences.
%
%METEOR is a precision-based metric for evaluating machine-translation out- put.
%
%BERTScore is a contextual embedding based metric for ...

Evaluated using BLEU \cite{papineni-etal-2002-bleu}, ROUGE \cite{lin-2004-rouge}, METEOR \cite{banerjee-lavie-2005-meteor} and BERTScore \cite{bert-score}, we find that the learning rates ranging from $10^{-4}$ to $10^{-5}$ for T5-Base and the learning rates ranging from $10^{-5}$ to $10^{-6}$ for T5-Large perform better.
%large model performed better results.
%
%After human evaluation, 
%Specifically, 
%we find that T5-Base with a learning rate of $3 \times 10^{-5}$ and T5-Large with a learning rate of $6 \times 10^{-6}$ produce the best results, and so we use these models to generate questions. 
Moreover, as expected,
the overall performance of T5-Large is better than T5-Base. 

Tables \ref{tab:auto-evaluation-base} and \ref{tab:auto-evaluation-large}
depict the measurement results for T5-Base and T5-Large, respectively, where R1, R2, RL, and RLsum stand for, respectively,
Rouge-1, Rouge-2, Rouge-L, and Rouge-Lsum. %The average is taken over all measures after multiplying the BERTScore scores by 100. 
The boldfaced number indicates the best in its column. It is evident that T5-Base with the
learning rate of $3 \times 10^{-5}$ and
T5-Large with the learning rate of $8 \times 10^{-6}$ produce the best results.
For convenience, we refer to these two finetuned models as T5-Base-SQuAD$_1$ and T5-Large-SQuAD$_1$
to distinguish them with the existing T5-Base-SQuAD model.
We sometimes also denote T5-Base-SQuAD$_1$ as T5-SQUAD$_1$ when there is no
confusion of what size of the dataset is used to pretrain T5.
% and so we use it to generate questions. 
%picked these models for our QG system.

\begin{table}[h]
\centering
\caption{Automatic Evaluation of T5-Base-SQuAD$_1$}
\begin{tabular}{l|c|c|c|c|c|c|c|c}
\hline
\textbf{Learning Rate} & \textbf{BLEU}  & \textbf{R1}    & \textbf{R2}    & \textbf{RL}    & \textbf{RLsum} & \textbf{METEOR} & \textbf{BERTScore} & \textbf{Average} \\ \hline
5e-5              & 20.01 & 50.71 & 28.38 & 46.59 & 46.61 & 45.46  & 51.51 & 41.32     \\ \hline
3e-5              & \textbf{22.63} & \textbf{54.90} & \textbf{32.22} & \textbf{50.97} & \textbf{50.99} & \textbf{48.98}  & \textbf{55.82} & \textbf{45.22}     \\ \hline
2.5e-5            & 22.50 & 54.36 & 31.93 & 50.49 & 50.50 & 48.64  & 55.61 & 44.86     \\ \hline
1e-5              & 20.17 & 50.46 & 28.38 & 46.79 & 46.81 & 44.97  & 51.82 & 41.34     \\ \hline
Dynamic %1.2e-4 $\rightarrow$ 3e-5 
& 20.57 & 51.88 & 28.99 & 47.67 & 47.68 & 47.38  & 53.34 & 42.50     \\ \hline
%Dynamic automatic      & 4.74  & 22.49 & 7.54  & 19.04 & 19.05 & 19.84  & 13.94 & 15.23     \\ \hline
\end{tabular}
\label{tab:auto-evaluation-base}
\end{table}

\begin{table}[h]
\centering
\caption{Automatic Evaluation on T5-Large-SQuAD$_1$}
\begin{tabular}{l|c|c|c|c|c|c|c|c}
\hline
\textbf{Learning Rate} & \textbf{BLEU}  & \textbf{R1}    & \textbf{R2}    & \textbf{RL}    & \textbf{RLsum} & \textbf{METEOR} & \textbf{BERTScore} & \textbf{Average} \\ \hline
3e-5                    & 23.01 & 54.49 & 31.92 & 50.51 & 50.51 & 50.00  & 56.19 & 45.23    \\ \hline
1e-5                    & 23.66 & 51.88 & 32.88 & 51.43 & 51.42 & 50.53  & 56.65 & 45.50   \\ \hline
8e-6                    & 23.83 & \textbf{55.48} & \textbf{33.08} & \textbf{51.58} & \textbf{51.58} & 50.61  & \textbf{56.94} & \textbf{46.15}   \\ \hline
6e-6                    & \textbf{23.84} & 55.24 & 32.91 & 51.35 & 51.35 & \textbf{50.70}  & 56.57 & 45.99   \\ \hline
Dynamic                 & 20.86 & 52.00 & 29.46 & 48.03 & 48.03 & 47.68  & 53.85 & 42.84   \\ \hline
\end{tabular}
\label{tab:auto-evaluation-large}
\end{table}

\section{Processing Pipelines} %Candidate Answers selecting and filtering}
\label{sec:P3}
%preprocessing

\begin{figure}[h]
  \centering
  \includegraphics[width= \linewidth]{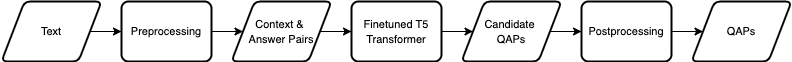}
  \caption{TP3 Architecture}
  \label{fig:2}
%   \Description{}
\end{figure}

The processing pipelines consist of preprocessing to select appropriate answers, question generation, and postprocessing to filter undesirable questions (see Fig. \ref{fig:2}).

\subsection{Preprocessing}

We observe that how to choose an answer would affect the quality of a question generated for the answer.
%however, not all answers can generate a satisfying question, 
%In general, T5 models work well on unseen data. and generates well-formed and grammatically correct questions from the given answer and context. There are, however, a few types of questions that could lead to generating incoherent questions. For example, asking about the type of verbs or asking about the clauses that express reasons or purposes may generate incoherent QAPs. 
We use a combination of NLP tools and algorithms to construct a preprocessing pipeline for
selecting appropriate answers as follows: % and filter undesirable questions. 
%
%In preprocessing stage, we select candidate answers by following 3 steps:
\begin{enumerate}
\item \textit{Remove unsuitable sentences}. We
first remove all interrogative and imperative sentences from the given article. We may do so 
by, for instance, simply removing any sentence that begins with a WH word or a verb and any sentence that ends with a question mark.
We then use semantic-role labeling \cite{Shi2019SimpleBM} to analyze sentences and remove those
that do not have any one of the following semantic-role tags: %ach sentence must consist of 3 semantic roles: 
subject, verb, and object. % we filtered out if any sentence doesn't satisfy this requirement. 
For each remaining sentence, if the total number of words contained in it, excluding stop words, is less than 4, then remove this sentence.
We then label the remaining sentences as \textit{suitable} sentences.

\item \textit{Remove candidate answers with inappropriate semantic-role labels}.
%The basic idea of candidate answers extraction is that, 
%We'd like to extract 
Nouns and phrasal nouns are candidate answers. But not any noun or phrasal noun would be suitable to be an answer.
We'd want a candidate answer to associate with a specific meaning. 
%In particular, we'd select candidate answer
%with a semantic-role label of subject, object, or manner. 
%To do so, 
%we first remove candidate answers with inappropriate semantic-role labels.
%We'd also like to extract 
%semantic roles with nouns. % as the main word in the role.
%To achieve this, 
Specifically, if a noun in a suitable sentence is identified as a name entity \cite{Peters2017SemisupervisedST}
or has a semantic-role label in the set of $\{$ARG, TMP, LOC, MNR, CAU, DIR$\}$, then keep it
as a candidate answer and remove the rest,
where ARG represents subject or object,
TMP represents time, LOC represents location, MNR represents manner, CAU represents cause, and DIR represents direction.
%
%with the following semantic-role labels as candidate answers and remove: ARG, TMP, LOC, MNR, and CAU as candidate answers, as well as words with named-entity labels \cite{Peters2017SemisupervisedST}. 
If a few candidate nouns occur consecutively, we treat the sequence of these nouns as a candidate answer phrase.

For example, in the sentence ``The engineers at the Massachusetts Institute of Technology (MIT) have taken it a step further changing the actual composition of plants in order to get them to perform diverse, even unusual functions", the phrase ``Massachusetts Institute of Technology" is recognized as a named entity, without a semantic-role label. Thus, it should not be selected as an answer. If it is selected, then the following QAP
(``Where is MIT located", ``Massachusetts Institute of Technology") will be generated, which is inadequate. 

\item \textit{Remove or prune answers with inadequate POS tags}.
Using semantic-role labels to identify what nouns to keep does not always work.
%works well most of the times, but
%there are situations that it would produce inadequate QAPs. 
%words are labeled wrong or labels redundant words. 
For example, the phrasal noun ``This widget" in the sentence ``This widget is more technologically advanced now"
has a semantic-role label of ARG1 (subject), which leads to the generation of the following question: 
``What widget is more technologically advanced now?" It is evident that this QAP is inadequate even though it is
grammatically correct. 
Note that ``This" has
a POS (part-of-speech) tag of PDT (predeterminer).
%This example indicates that semantic-role labeling alone is insufficient and POS tagging should also be used.
%is not necessary for the answer and it is labeled as predeterminer (PDT) by POS tagging, if the model generates question according to this answer phrase, it may generate inappropriate question "What widget is more technologically advanced now?" with answer "This widget".
For another example, while the word ``now" in the sentence 
has a semantic-role label of TMP (time), its POS tag is RB (adverb).
In general, we remove nouns with a POS tag in $\{$RB, RP, CC, DT, IN, MD, PDT, PRP, WP, WDT, WRB$\}$ 
or prune words with such a POS tag at either end of a phrasal noun.
%
%should be removed. In general,
%, it should not be used as answer for question generation.
%we remove such answers or keep the noun component in a phrase by pruning the words at either end 
%with %in an answer phrase according to 
%the following POS tags: 
%RB, RP, CC, DT, IN, MD, PDT, PRP, WP, WDT, WRB. 
After this treatment, the candidate answer ``now" is removed and the candidate answer phrase ``This widget" is pruned to ``widget". For this answer and the input sentence, the following question is generated: ``What is more technologically advanced now?" Evidently this question is more adequate.

\item \textit{Remove common answers}. %%%needs work
%Through experiments, % and generalization, 
We observe that certain candidate answers, such as ``anyone", ``people", and ``stuff", 
would often lead to generation of inadequate questions. 
%For example, "anyone", "people", "stuff" etc. 
Such words tend to be common words that should be removed. 
We do so by looking up the probabilities of 1-grams from the
%language model of the 
Google Books Ngram Dataset \cite{doi:10.1126/science.1199644}. If the probability of a noun word is greater 
than 0.15\%, we remove its candidacy. Likewise, we may also treat noun phrases by looking up the probabilities $n$-grams for $n > 1$, but doing so would incur much more processing time.

\item \textit{Filtering answers appearing in clauses}.
We observe that a candidate answer appearing in the latter part of a clause would often lead to a generation of an
inadequate QAP. Such candidate answers would appear at lower levels in a dependency tree. We use the following procedure to identify such candidate answers: For each remaining sentence $s$, we first generate its dependency tree \cite{varga2010wlv}. Let $h_s$ be the height of the tree. Suppose that a candidate answer $a$ appears in a clause contained in $s$. If $a$ is a single noun, let its height in the tree be $h_a$. If $a$ is a phrasal noun, 
let the average height of the heights of the words contained in $a$ be $h_a$. If $h_a \geq \tfrac{2}{3}h_s$, then remove $a$.

Take the following sentence as an example: ``While I tend to buy a lot of books, these three were given to me as gifts, which might add to the meaning I attach to them." In 
this sentences, the following noun ``gifts” and phrasal nouns ``a lot of books" and ``the meaning I attach to them"are labeled as object.
However, T5
% is labeled as object, where 
%``a lot of books" and ``gifts" both reference the same semantic-role label of object and the underlying transformer may not be able to
resolves multiple objects poorly, 
and if we choose ``the meaning I attach to them" as an answer, T5 will generate the following question: 
%answer phrase "the meaning I attach to them" is labeled as object in " which" clause in the sentence , since  and the model sometime can't resolve multiple objects appropriately, the phrase may leads to generate an inadequate question 
``What did the gifts add to the books", which is inadequate. Since this phrasal noun appears in a clause and at a lower level of the dependency tree, it is removed from being selected as a candidate answer.

%and its height is at least 2/3 of the height of the tree, then remove it.
%is higher than that of the verb node, 
%and if it higher than 75\% of the dependency tree height, we filter the it out, 
%If the answer is a phrasal noun, we take the average height of each word in the phrase as its height. 

\item \textit{Removing redundant answers}.
%We remove duplicate answers.
% by Named Entity labeling and semantic-role labeling, 
%If a candidate answer is contained in a longer answer phrase, we remove the shorter one.
If a candidate answer word or phrase is contained in another candidate answer phrase and appear in the same sentence,
we extract from the dependency tree of the sentence the subtree $T_s$ for the shorter candidate phrase and subtree $T_l$ for the longer candidate phrase, then $T_s$ is also a subtree of $T_l$. If $T_s$ and $T_l$ share the same root, %is same as the root part of the $h_l$, we say 
then the shorter candidate answer is more syntactically important than the longer one, and so we remove the longer candidate answer. Otherwise, remove the shorter candidate answer.
 
Take the sentence ``The longest track and field event at the Summer Olympics is the 50-kilometer race walk, which is about five miles longer than the marathon" as an example.  
The shorter phrase ``Summer Olympics" is recognized as a named entity, 
%and a longer phrase "The longest track and field event at the Summer Olympics" is labeled as subject by semantic role.
which leads to the generation of the following inadequate QAP: (``What is the longest track and field event", ``Summer Olympics). 
On the other hand,
the longer phrase ``The longest track and field event at the Summer Olympics" is labeled as subject for its semantic role, which leads to the generation of the following adequate QAP: (``What is the 50-kilometer race walk", ``The longest track and field event at the Summer Olympics").
Since the root word for the longer phrase is ``event" that is not contained in the shorter phrase, so the shorter phrase is removed to avoid generating the inadequate QAP.

\end{enumerate}

\subsection{Question generation}

After extracting all candidate answers from the preprocessing pipeline, for each answer extracted, we use three adjacent sentences as the context, with the middle sentence containing the answer, and concatenate the answer and the context with marks into the following format as input to a fine-turned T5 model: $\langle answer\rangle$answer\_text$\langle context\rangle$context\_text, to generate candidate questions.
We note that the greedy search in the decoder of the T5 model does not guarantee the optimal result, we use beam search with 3 beams to select the word sequences with the top 3 probabilities from the probability distribution and acquire 3 candidate questions.
We then concatenate each candidate question with the corresponding answer as a new sentence and generate
an embedding vector representation for it using %map the sentence to %dense vector space 
%sentence embeddings by utilizing 
the pretrained  RoBERTa-Large model \cite{Liu2019RoBERTaAR,reimers-2019-sentence-bert}, and select the most semantically similar question to the context as the final target question.

\subsection{Postprocessing}

Recall that in the preprocessing pipeline, %the answers selecting and filtering stage, 
we have removed inappropriate candidate answers.
% that would lead to generating appropriate questions.
However, some of the remaining answers may still lead to generating inappropriate questions. 
Thus, in the postprocessing pipeline, we proceed to remove inadequate questions as follows:
%by following 3 steps to tackle this problem:
\begin{enumerate}
\item \textit{Remove questions that contain the answers}.
Remove a question if the corresponding answer or the main body of the answer is contained in the question.
%\hl{Changes start here}
If the answer includes a clause, we extract the main body of the answer as follows:
Parse the answer to constituency tree \cite{Joshi2018ExtendingAP} and remove the subtree rooted with a subordinate clause label SBAR, the remaining part of the phrase is the main body of the answer.

For example, in the sentence ``The first, which I take to reading every spring is Ernest Hemningway's A Moveable Feast", ``The first, which I take to reading every spring" is labeled as subject. Using it as a candidate answer  generates an inadequate question for the answer ``What is the first book I reread?" Note that the phrase ``The first" can be extracted as the main body of the answer, which is contained in the question. Thus, this QAP is removed.
%the generated question can be removed.
%\hl{Changes end here}

\item \textit{Remove short questions}.
If the generated question, after removing stop words, consists of only one word, then remove the question.
%we filter this question. 
For example, ``What is it?" and ``Who is she?" will be removed because after removing stop words,
the former becomes ``What" and the latter becomes ``Who". On the other hand, ``Where is Boston?" will remain.

\item \textit{Remove unsuitable questions}.
Recall that we generate the question from the adjacent three sentences in the article, with the middle sentence containing the answer. However, the middle sentence may not be the only sentence containing the answer. In other words,
the first or the last sentences may also contain the answer. 
Assuming that all three sentences contain the answer, our finetuned T5 transformer may generate a question based on the first sentence or the last sentence. If the first sentence or the last sentence is not a suitable sentence
we labeled in the preprocessing pipeline, 
%our generatable sentence list which we described in the preprocessing stage, 
the question being generated may be in appropriate. 
We'd want to make sure that the question is generated for a suitable sentence.
For this purpose, we first identify which sentence the question is generated for. In particular,
%it may cause generating inappropriate questions.
let $s_i$ for $i=1,2,3$ be the 3 sentences and $(q,a)$ be the question generated for answer $a$.
Let $QA$ denote the union of the set of words in $q$ and the set of words in $a$.
Likewise, let $S_i$ be the set of words in $s_i$. If $QA \cap S_i$ is the largest among the other two
intersections, then $q$ is likely generated from $s_i$ for $a$. If $s_i$ is not suitable, then
remove $q$.

Note that we may also consider word sequences in addition to word sets. For example, we may consider longest common subsequences or longest common substrings when comparing two word sequences. But in our experiments, they don't seem to 
add extra benefits.
%a word sequence that concatenate the word sequence in the generated question 
%and the word sequence in the answer, and form three other word sequences each of which corresponds to an input sentence.
%
%in the order they appear and the answer into a word sequence, and place words in each sentence in the order they appear into another word sequence.
%We then count the number of words in the intersection of two word sequences and the number of longest common subsequence for question answer with each sentence. If the sentence has more intersection and more longest common subsequence with the question and answer, we think the question is generated from that sentence, and if the sentence is not in the generatable sentence list, we filter out the corresponding generated question.

%(4) Prune question,  parser question to cp tree,  if question contains clause (tree node contains SBAR), and the question longer than 10 words, remove the leaves followed by when, where, which. 
\end{enumerate}

\section{Evaluations}\label{sec:evaluations}

To evaluate the quality of QAPs generated by TP3-Base and TP3-Large, we use the standard automatic evaluation metrics as well as
human judgments.

\subsection{Automatic evaluations}

We first compare T5-SQuAD$_1$ with the exiting QG models 
with the standard automatic evaluation metrics as before:
BLEU, % \cite{10.3115/1073083.1073135}, 
ROUGE-1 (R1), ROUGE-2 (R2), ROUGE-L (RL), ROUGE-LSum (RLsum), % \cite{lin-2004-rouge}, 
METEOR (MTR), % \cite{banerjee-lavie-2005-meteor}) to automatically evaluate our finetuned models, those n-grams-based metrics was used to evaluate the syntactic reconstruction ability of the models. Besides the n-grams-based metrics, we also used BERTScore \cite{bert-score} as semantic-based metric to evaluate the semantic reconstruction ability of the models.
and BERTScore (BScore). 
Since most existing QG models are based on pretrained transformers with the base dataset, we will
compare T5-Base-SQuAD$_1$ with the existing QG models. 
%The results are shown on
%Table \ref{tab:auto-evaluation}.

\begin{table}[h]
\centering
\caption{Automatic evaluation results}
\begin{tabular}{l|c|c|c|c|c|c|c|c|c}
\hline
\textbf{Model}  & \textbf{Size}  & \textbf{BLEU} & \textbf{R1}  & \textbf{R2} & \textbf{RL} & \textbf{RLsum}          & \textbf{MTR}    & \textbf{BScore}     &\textbf{Average}  \\ \hline
ProphetNet  & Large & 22.88          & 51.37          & 29.48          & 47.11          & 47.09          & 41.46          & 49.31     & 41.24   \\ \hline
BART-hl     & Base  & 21.13          & 51.88          & 29.43          & 48.00          & 48.01          & 40.23          & 54.33     & 41.86    \\ \hline
BART-SQuAD  & Base  & 22.09          & 52.75          & 30.56          & 48.79          & 48.78          & 41.39          & 54.86     & 42.75    \\ \hline
T5-hl       & Base  & 23.19          & 53.52          & 31.22          & 49.40          & 49.40          & 42.68          & 55.48     & 43.56    \\ \hline
T5-SQuAD    & Base  & \textbf{23.74} & 54.12          & 31.84          & 49.82          & 49.81          & 43.63          & 55.68     & 44.09    \\ \hline
MixQG$_1$       & Base  & 23.53          & 54.39          & 32.06          & 50.05          & 50.02          & 43.83          & 55.66     & 44.22    \\ \hline
MixQG$_2$       & Base  & 23.74          & 54.28          & 32.23          & 50.35          & 50.34          & 43.91          & 55.71     & 44.37    \\ \hline
MixQG-SQuAD & Base  & 23.46          & 54.48          & 32.18          & 50.14          & 50.10          & 44.15          & \textbf{55.82} & 44.33\\ \hline
T5-SQuAD$_1$ & Base  & 22.62          & \textbf{54.87} & \textbf{32.20} & \textbf{50.99} & \textbf{50.98} & \textbf{48.98} & \textbf{55.82} & \textbf{45.21}  \\ \hline
\end{tabular}
\label{tab:auto-evaluation}
\end{table}

Table \ref{tab:auto-evaluation} shows automatic evaluation comparison results with 
%ProphetNet \cite{qi2020prophetnet}, BART-HL and BART-SQuAD \cite{lewis-etal-2020-bart}; T5-HL and T5 \cite{raffel2020exploring}; and MixQG$_1$, MixQG$_2$, and MixQG-SQuAD \cite{murakhovska2021mixqg}. All but ProphetNet are pretrained on the base dataset and then finetuned on the SQuAD dataset. 
%\hl{ Changes start here: }
ProphetNet \cite{qi2020prophetnet}, BART \cite{lewis-etal-2020-bart}, T5 \cite{raffel2020exploring} and MixQG \cite{murakhovska2021mixqg}. 
BART-SQuAD, T5-SQuAD, and MixQG-SQuAD are corresponding models finetuned on the SQuAD dataset.
BART-hl and T5-hl are augmented models using the ``highlight" encoding scheme introduced by Chan and Fan \cite{chan2019recurrent}.
%\hl{ Changes end here}

The results of MixQG$_1$ were presented in the original paper \cite{murakhovska2021mixqg}, and the results of MixQG$_2$ were computed by us using the pretrained model posted on HuggingFace (https://huggingface.co/Salesforce/mixqg-base). %The result of MetaQG was our improved model.
The results show that, except BLEU, T5-SQuAD$_1$ outperforms all other models on the ROUGE and METEOR metrics,  produces the same BERTScore score as that of MixQG-SQuAD. 
%T5-SQuAD$_1$'s BLEU score is slightly lower than MixQG because some of the lengths of the generated questions are shorter which results in Brevity Penalty in the BLEU algorithm. 
 Overall, T5-SQuAD$_1$ performs better than all the models in comparison.

\subsection{Manual evaluations of TP3}

A number of publications (e.g., see \cite{callison-burch-etal-2006-evaluating,liu-etal-2016-evaluate,nema-khapra-2018-towards}) have shown that the aforementioned automatic evaluation metrics based on n-gram similarities do not always 
correlate well with human judgments about the answerability of a question. 
Thus, we'd also need to use human experts to evaluate the qualities of QAPs generated
by TP3. We do so 
%We then manually evaluated our base and large models 
on the Gaokao-EN dataset consisting of 85 articles, where each article contains 15 to 20 sentences. We chose Gaokao-EN because expert evaluations are provided to us from a project we work on.
%TP3 generates a total of at least 1,270 QAPs. 
Table \ref{tab:manually-evaluation} depicts the evaluation results. Title abbreviations are explained below,
where the numbers in boldface are the best in the corresponding columns:
\begin{enumerate}
\item \textbf{Total} means the total number of QAPs generated by TP3. 
\item \textbf{ADQT} means the total number of adequate QAPs. These QAPs can be directly used without any modification. 
\item \textbf{ACPT} means the total number of
QAPs where the question, while semantically correct, contains a minor English issue that can be corrected with a minor effort. For example,
a question may simply be missing a word or a phrase at the end. Such QAPs may be deemed acceptable. 
\item \textbf{UA} means unacceptable QAPs. 
\item \textbf{ADQT-R} means the ratio of the adequate QAPs over all the QAPs being generated.
\item \textbf{ACPT-R} means the ratio of the adequate and acceptable QAPs over all the QAPs being generated.
%\item \textbf{Better} means the QAPs among the adequate QAPs under the current model that are better than the adequate QAPs generated under different models with respect to the same answers.
\end{enumerate}

\begin{table}[h]
\centering
\caption{Manual evaluation results for TP3-Base and TP3-Large over Gaokao-EN}
\begin{tabular}{l|c|c|c|c|c|c|c}
\hline
\textbf{TP3} &	\textbf{Learning Rate} &	\textbf{Total} & \textbf{ADQT} & \textbf{ACPT}	 & \textbf{UA} 
	& \textbf{ADQT-R}	& \textbf{ACPT-R} %& \textbf{Better} 	
	\\ \hline
Base		           & 3e-5 & \textbf{1290}& 1145& \textbf{63}& 82& 88.76& 93.64 %& 20
\\ \hline
\multirow{5}{*}{Large} & 3e-5 & 1287& 1162& 49& 76& 90.29& 94.09 %& 19
\\\cline{2-8}
   					   & 1e-5 & 1271& 1166& 39& 76& 91.74& 94.81 %& 16
   					   \\\cline{2-8}
   					   & 8e-6 & 1270& 1162& 39& 69& 91.50& 94.57 %& 27
   					   \\\cline{2-8}
   					   & 6e-6 & 1273& \textbf{1172}& 45& \textbf{56}& \textbf{92.07}& \textbf{95.60} %& \textbf{36}
   					   \\\cline{2-8}
   				 & Dynamic & 1288& 1116& 51& 121& 86.65& 90.61 %& 22
   				 \\\hline
\end{tabular}
\label{tab:manually-evaluation}
\end{table}

%The performance of T5 base models with different learning rate are shown in Table \ref{tab:base} and the T5 Large models are shown in Table \ref{tab:large}, The base model with learning rate $3 \times 10^{-5}$ and the large model with learning rate $6 \times 10^{-6}$ generated less non-acceptable questions, so we picked these models for our QG system.

%\begin{table}[]
%\begin{tabular}{|l|l|l|l|l|l|l|}
%\hline
%Learning Rate  & 1e-4 & 7.5e-5 & 5e-5 & 2.5e-5 & 1e-5 & Dynamic LR \\ \hline
%Good           &      &      &        &      &        &      \\ \hline
%Acceptable     &      &      &        &      &        &      \\ \hline
%Not acceptable &      &      &        &      &        &      \\ \hline
%\end{tabular}
%\caption{T5 base model evaluated on 300 questions}
%\label{tab:base}
%\end{table}
%
%\begin{table}[]
%\begin{tabular}{|l|l|l|l|l|l|l|}
%\hline
%Learning Rate  & 3e-5 & 1e-5 & 8e-6   & 6e-6 & Dynamic LR   \\ \hline
%Good           &      &      &        &      &              \\ \hline
%Acceptable     &      &      &        &      &              \\ \hline
%Not acceptable &      &      &        &      &              \\ \hline
%\end{tabular}
%\caption{T5 large model evaluated on 500 questions}
%\label{tab:large}
%\end{table}

\section{Conclusions}\label{sec:conclusions}

We presented a downstream task of transformers on generating question-answer pairs by finetuning pretrained T5 models with preprocessing and postprocess pipelines, and generate a satisfactory number of adequate QAPs for a given article with high qualities. To facilitate reproduction and further investigation, we have released the source code at https://github.com/zhangchengx/T5-Fine-Tuning-for-Question-Generation and the model at https://huggingface.co/ZhangCheng/T5-Base-finetuned-for-Question-Generation. The Gaokao-EN dataset and the
human judgments of QAPs are also available at https://github.com/zhangchengx/Gaokao-EN

%We noted that properly selecting a pretrained transformer model and finetuning it is essential. 
With an improved 
transformer it is possible to improve both the number and qualities of QAPs being generated. It's also possible to strengthen the preprocessing and postprocessing pipelines. For example, in addition to using a 1-gram language model to determine if a candidate answer would be appropriate, we may develop a more efficient method to use $n$-gram language models for checking a candidate answer being a phrasal noun. Also, when we feed a context to a transformer, in addition to feeding the model with three consecutive sentences in the article as we currently do, there are
other ways to select sentences. For example, we may consider clustering similar sentences and rank them three at a time, such that the sentence in the middle contains the selected candidate answer.
Another direction would be to explore how to generate QAPs for candidate answers that appear at lower levels of dependency trees. 
These issues deserve further investigations.
%In this paper, we present a T5 Transformer question generation model finetuned on SQuQD datasets with preprocessing and postprocessing pipelines to select appropriate answers and filter the undesirable questions. 
%Through experiments and evaluations we show that our model outperforms other compared models on most metrics, it not only achieves state-of-the-art results on target datasets but also achieves satisfactory results on human evaluation.
%We also release our source code (https://github.com/zhangchengx/T5-Fine-Tuning-for-Question-Generation) and the model (https://huggingface.co/ZhangCheng/T5-Base-finetuned-for-Question-Generation) to facilitate reproducibility and future work.
%In this paper, we present a T5 Transformer question generation model finetuned on SQuQD datasets with preprocessing and postprocessing pipelines to select appropriate answers and filter the undesirable questions. 
%Through experiments and evaluations we show that our model outperforms other compared models on most metrics, it not only achieves state-of-the-art results on target datasets but also achieves satisfactory results on human evaluation.
%We also release our source code (https://github.com/zhangchengx/T5-Fine-Tuning-for-Question-Generation) and the model (https://huggingface.co/ZhangCheng/T5-Base-finetuned-for-Question-Generation), and Gaokao data (https://github.com/zhangchengx/Gaokao-Articles) to facilitate reproducibility and future work.

\bibliographystyle{apalike}

\bibliography{bibliography}

\begin{thebibliography}{}

\bibitem[Ali et~al., 2010]{ali2010automation}
Ali, H., Chali, Y., and Hasan, S.~A. (2010).
\newblock Automation of question generation from sentences.
\newblock In {\em Proceedings of QG2010: The Third Workshop on Question
  Generation}, pages 58--67.

\bibitem[Bahdanau et~al., 2014]{bahdanau2014neural}
Bahdanau, D., Cho, K., and Bengio, Y. (2014).
\newblock Neural machine translation by jointly learning to align and
  translate.
\newblock {\em arXiv preprint arXiv:1409.0473}.

\bibitem[Banerjee and Lavie, 2005]{banerjee-lavie-2005-meteor}
Banerjee, S. and Lavie, A. (2005).
\newblock {METEOR}: an automatic metric for {MT} evaluation with improved
  correlation with human judgments.
\newblock In {\em Proceedings of the {ACL} Workshop on Intrinsic and Extrinsic
  Evaluation Measures for Machine Translation and/or Summarization}, pages
  65--72, Ann Arbor, Michigan. Association for Computational Linguistics.

\bibitem[Callison-Burch et~al., 2006]{callison-burch-etal-2006-evaluating}
Callison-Burch, C., Osborne, M., and Koehn, P. (2006).
\newblock Re-evaluating the role of {B}leu in machine translation research.
\newblock In {\em 11th Conference of the {E}uropean Chapter of the Association
  for Computational Linguistics}, pages 249--256, Trento, Italy. Association
  for Computational Linguistics.

\bibitem[Chan and Fan, 2019]{chan2019recurrent}
Chan, Y.-H. and Fan, Y.-C. (2019).
\newblock A recurrent {BERT}-based model for question generation.
\newblock In {\em Proceedings of the 2nd Workshop on Machine Reading for
  Question Answering}, pages 154--162.

\bibitem[Chen et~al., 2020]{Chen2020ReinforcementLB}
Chen, Y., Wu, L., and Zaki, M.~J. (2020).
\newblock Reinforcement learning based graph-to-sequence model for natural
  question generation.
\newblock {\em ArXiv}, abs/1908.04942.

\bibitem[Cho et~al., 2014]{cho-etal-2014-learning}
Cho, K., van Merri{\"e}nboer, B., Gulcehre, C., Bahdanau, D., Bougares, F.,
  Schwenk, H., and Bengio, Y. (2014).
\newblock Learning phrase representations using {RNN} encoder{--}decoder for
  statistical machine translation.
\newblock In {\em Proceedings of the 2014 Conference on Empirical Methods in
  Natural Language Processing ({EMNLP})}, pages 1724--1734, Doha, Qatar.
  Association for Computational Linguistics.

\bibitem[Danon and Last, 2017]{danon2017syntactic}
Danon, G. and Last, M. (2017).
\newblock A syntactic approach to domain-specific automatic question
  generation.
\newblock {\em arXiv preprint arXiv:1712.09827}.

\bibitem[Dong et~al., 2019]{dong2019unified}
Dong, L., Yang, N., Wang, W., Wei, F., Liu, X., Wang, Y., Gao, J., Zhou, M.,
  and Hon, H.-W. (2019).
\newblock Unified language model pre-training for natural language
  understanding and generation.
\newblock In {\em 33rd Conference on Neural Information Processing Systems
  (NeurIPS 2019)}.

\bibitem[Du et~al., 2017]{du-etal-2017-learning}
Du, X., Shao, J., and Cardie, C. (2017).
\newblock Learning to ask: neural question generation for reading
  comprehension.
\newblock In {\em Proceedings of the 55th Annual Meeting of the Association for
  Computational Linguistics (Volume 1: Long Papers)}, pages 1342--1352,
  Vancouver, Canada. Association for Computational Linguistics.

\bibitem[Duan et~al., 2017]{duan-etal-2017-question}
Duan, N., Tang, D., Chen, P., and Zhou, M. (2017).
\newblock Question generation for question answering.
\newblock In {\em Proceedings of the 2017 Conference on Empirical Methods in
  Natural Language Processing}, pages 866--874, Copenhagen, Denmark.
  Association for Computational Linguistics.

\bibitem[Harrison and Walker, 2018]{Harrison_2018}
Harrison, V. and Walker, M. (2018).
\newblock Neural generation of diverse questions using answer focus, contextual
  and linguistic features.
\newblock In {\em Proceedings of the 11th International Conference on Natural
  Language Generation}. Association for Computational Linguistics.

\bibitem[Heilman and Smith, 2009]{heilman2009question}
Heilman, M. and Smith, N.~A. (2009).
\newblock Question generation via overgenerating transformations and ranking.
\newblock Technical report, CARNEGIE-MELLON UNIV PITTSBURGH PA LANGUAGE
  TECHNOLOGIES INST.

\bibitem[Joshi et~al., 2018]{Joshi2018ExtendingAP}
Joshi, V., Peters, M.~E., and Hopkins, M. (2018).
\newblock Extending a parser to distant domains using a few dozen partially
  annotated examples.
\newblock In {\em ACL}.

\bibitem[Khullar et~al., 2018]{khullar2018automatic}
Khullar, P., Rachna, K., Hase, M., and Shrivastava, M. (2018).
\newblock Automatic question generation using relative pronouns and adverbs.
\newblock In {\em Proceedings of ACL 2018, Student Research Workshop}, pages
  153--158.

\bibitem[Kim et~al., 2019]{Kim_2019}
Kim, Y., Lee, H., Shin, J., and Jung, K. (2019).
\newblock Improving neural question generation using answer separation.
\newblock In {\em Association for the Advancement of Artificial Intelligence
  (AAAI)}, volume~33, page 6602–6609. Association for the Advancement of
  Artificial Intelligence (AAAI).

\bibitem[Kriangchaivech and Wangperawong, 2019]{Kriangchaivech190905017}
Kriangchaivech, K. and Wangperawong, A. (2019).
\newblock Question generation by transformers.

\bibitem[Labutov et~al., 2015]{labutov-etal-2015-deep}
Labutov, I., Basu, S., and Vanderwende, L. (2015).
\newblock Deep questions without deep understanding.
\newblock In {\em Proceedings of the 53rd Annual Meeting of the Association for
  Computational Linguistics and the 7th International Joint Conference on
  Natural Language Processing (Volume 1: Long Papers)}, pages 889--898,
  Beijing, China. Association for Computational Linguistics.

\bibitem[Lai et~al., 2017]{lai2017large}
Lai, G., Xie, Q., Liu, H., Yang, Y., and Hovy, E. (2017).
\newblock Race: Large-scale reading comprehension dataset from examinations.
\newblock {\em arXiv preprint arXiv:1704.04683}.

\bibitem[Lelkes et~al., 2021]{10.1145/3442381.3449892}
Lelkes, A.~D., Tran, V.~Q., and Yu, C. (2021).
\newblock Quiz-style question generation for news stories.
\newblock In {\em Proceedings of the Web Conference 2021}, WWW '21, page
  2501–2511, New York, NY, USA. Association for Computing Machinery.

\bibitem[Lewis et~al., 2020]{lewis-etal-2020-bart}
Lewis, M., Liu, Y., Goyal, N., Ghazvininejad, M., Mohamed, A., Levy, O.,
  Stoyanov, V., and Zettlemoyer, L. (2020).
\newblock {BART}: Denoising sequence-to-sequence pre-training for natural
  language generation, translation, and comprehension.
\newblock In {\em Proceedings of the 58th Annual Meeting of the Association for
  Computational Linguistics}, pages 7871--7880, Online. Association for
  Computational Linguistics.

\bibitem[Lin, 2004]{lin-2004-rouge}
Lin, C.-Y. (2004).
\newblock {ROUGE}: a package for automatic evaluation of summaries.
\newblock In {\em Text Summarization Branches Out}, pages 74--81, Barcelona,
  Spain. Association for Computational Linguistics.

\bibitem[Lindberg, 2013]{lindberg2013automatic}
Lindberg, D.~L. (2013).
\newblock {\em Automatic question generation from text for self-directed
  learning}.
\newblock PhD thesis, Applied Sciences: School of Computing Science.

\bibitem[Liu et~al., 2016]{liu-etal-2016-evaluate}
Liu, C.-W., Lowe, R., Serban, I., Noseworthy, M., Charlin, L., and Pineau, J.
  (2016).
\newblock How {NOT} to evaluate your dialogue system: an empirical study of
  unsupervised evaluation metrics for dialogue response generation.
\newblock In {\em Proceedings of the 2016 Conference on Empirical Methods in
  Natural Language Processing}, pages 2122--2132, Austin, Texas. Association
  for Computational Linguistics.

\bibitem[Liu et~al., 2019]{Liu2019RoBERTaAR}
Liu, Y., Ott, M., Goyal, N., Du, J., Joshi, M., Chen, D., Levy, O., Lewis, M.,
  Zettlemoyer, L., and Stoyanov, V. (2019).
\newblock Roberta: A robustly optimized bert pretraining approach.
\newblock {\em ArXiv}, abs/1907.11692.

\bibitem[Lopez et~al., 2020]{Lopez2020TransformerbasedEQ}
Lopez, L.~E., Cruz, D.~K., Cruz, J. C.~B., and Cheng, C.~K. (2020).
\newblock Transformer-based end-to-end question generation.
\newblock {\em ArXiv}, abs/2005.01107.

\bibitem[Loshchilov and Hutter, 2019]{loshchilov2018decoupled}
Loshchilov, I. and Hutter, F. (2019).
\newblock Decoupled weight decay regularization.
\newblock In {\em International Conference on Learning Representations}.

\bibitem[Luong et~al., 2015]{luong-etal-2015-effective}
Luong, T., Pham, H., and Manning, C.~D. (2015).
\newblock Effective approaches to attention-based neural machine translation.
\newblock In {\em Proceedings of the 2015 Conference on Empirical Methods in
  Natural Language Processing}, pages 1412--1421, Lisbon, Portugal. Association
  for Computational Linguistics.

\bibitem[Michel et~al., 2011]{doi:10.1126/science.1199644}
Michel, J.-B., Shen, Y.~K., Aiden, A.~P., Veres, A., Gray, M.~K., null null,
  Pickett, J.~P., Hoiberg, D., Clancy, D., Norvig, P., Orwant, J., Pinker, S.,
  Nowak, M.~A., and Aiden, E.~L. (2011).
\newblock Quantitative analysis of culture using millions of digitized books.
\newblock {\em Science}, 331(6014):176--182.

\bibitem[Murakhovs'ka et~al., 2021a]{mixqg.2110.08175}
Murakhovs'ka, L., Wu, C.-S., Niu, T., Liu, W., and Xiong, C. (2021a).
\newblock Mixqg: Neural question generation with mixed answer types.

\bibitem[Murakhovs'ka et~al., 2021b]{murakhovska2021mixqg}
Murakhovs'ka, L., Wu, C.-S., Niu, T., Liu, W., and Xiong, C. (2021b).
\newblock Mixqg: Neural question generation with mixed answer types.

\bibitem[Nema and Khapra, 2018]{nema-khapra-2018-towards}
Nema, P. and Khapra, M.~M. (2018).
\newblock Towards a better metric for evaluating question generation systems.
\newblock In {\em Proceedings of the 2018 Conference on Empirical Methods in
  Natural Language Processing}, pages 3950--3959, Brussels, Belgium.
  Association for Computational Linguistics.

\bibitem[Papineni et~al., 2002a]{10.3115/1073083.1073135}
Papineni, K., Roukos, S., Ward, T., and Zhu, W.-J. (2002a).
\newblock {BLEU}: a method for automatic evaluation of machine translation.
\newblock In {\em Proceedings of the 40th Annual Meeting on Association for
  Computational Linguistics}, ACL’02, page 311–318, USA. Association for
  Computational Linguistics.

\bibitem[Papineni et~al., 2002b]{papineni-etal-2002-bleu}
Papineni, K., Roukos, S., Ward, T., and Zhu, W.-J. (2002b).
\newblock {B}leu: a method for automatic evaluation of machine translation.
\newblock In {\em Proceedings of the 40th Annual Meeting of the Association for
  Computational Linguistics}, pages 311--318, Philadelphia, Pennsylvania, USA.
  Association for Computational Linguistics.

\bibitem[Peters et~al., 2017]{Peters2017SemisupervisedST}
Peters, M.~E., Ammar, W., Bhagavatula, C., and Power, R. (2017).
\newblock Semi-supervised sequence tagging with bidirectional language models.
\newblock In {\em ACL}.

\bibitem[Qi et~al., 2020a]{qi2020prophetnet}
Qi, W., Yan, Y., Gong, Y., Liu, D., Duan, N., Chen, J., Zhang, R., and Zhou, M.
  (2020a).
\newblock Prophetnet: Predicting future n-gram for sequence-to-sequence
  pre-training.
\newblock In {\em Proceedings of the 2020 Conference on Empirical Methods in
  Natural Language Processing: Findings}, pages 2401--2410.

\bibitem[Qi et~al., 2020b]{qi-etal-2020-prophetnet}
Qi, W., Yan, Y., Gong, Y., Liu, D., Duan, N., Chen, J., Zhang, R., and Zhou, M.
  (2020b).
\newblock {P}rophet{N}et: Predicting future n-gram for
  sequence-to-{S}equence{P}re-training.
\newblock In {\em Findings of the Association for Computational Linguistics:
  EMNLP 2020}, pages 2401--2410, Online. Association for Computational
  Linguistics.

\bibitem[Radford et~al., 2019]{radford2019language}
Radford, A., Wu, J., Child, R., Luan, D., Amodei, D., Sutskever, I., et~al.
  (2019).
\newblock Language models are unsupervised multitask learners.
\newblock {\em OpenAI blog}, 1(8):9.

\bibitem[Raffel et~al., 2020]{raffel2020exploring}
Raffel, C., Shazeer, N., Roberts, A., Lee, K., Narang, S., Matena, M., Zhou,
  Y., Li, W., and Liu, P.~J. (2020).
\newblock Exploring the limits of transfer learning with a unified text-to-text
  transformer.

\bibitem[Rajpurkar et~al., 2016]{Rajpurkar_2016}
Rajpurkar, P., Zhang, J., Lopyrev, K., and Liang, P. (2016).
\newblock Squad: 100,000+ questions for machine comprehension of text.
\newblock In {\em Proceedings of the 2016 Conference on Empirical Methods in
  Natural Language Processing}, pages 2383--2392. Association for Computational
  Linguistics.

\bibitem[Reddy et~al., 2019]{reddy2019coqa}
Reddy, S., Chen, D., and Manning, C.~D. (2019).
\newblock Coqa: A conversational question answering challenge.

\bibitem[Reimers and Gurevych, 2019]{reimers-2019-sentence-bert}
Reimers, N. and Gurevych, I. (2019).
\newblock Sentence-bert: Sentence embeddings using siamese bert-networks.
\newblock In {\em Proceedings of the 2019 Conference on Empirical Methods in
  Natural Language Processing}. Association for Computational Linguistics.

\bibitem[Sachan and Xing, 2018]{sachan-xing-2018-self}
Sachan, M. and Xing, E. (2018).
\newblock Self-training for jointly learning to ask and nnswer questions.
\newblock In {\em Proceedings of the 2018 Conference of the North {A}merican
  Chapter of the Association for Computational Linguistics: Human Language
  Technologies, Volume 1 (Long Papers)}, pages 629--640, New Orleans,
  Louisiana. Association for Computational Linguistics.

\bibitem[Sak et~al., 2014]{Sak2014LongSM}
Sak, H., Senior, A.~W., and Beaufays, F. (2014).
\newblock Long short-term memory recurrent neural network architectures for
  large scale acoustic modeling.
\newblock In {\em Proccedings of the 15th Annual Conference of the
  International Speech Communication Association}.

\bibitem[Shi and Lin, 2019]{Shi2019SimpleBM}
Shi, P. and Lin, J.~J. (2019).
\newblock Simple bert models for relation extraction and semantic role
  labeling.
\newblock {\em ArXiv}, abs/1904.05255.

\bibitem[Smith, 2017]{7926641}
Smith, L.~N. (2017).
\newblock Cyclical learning rates for training neural networks.
\newblock In {\em 2017 IEEE Winter Conference on Applications of Computer
  Vision (WACV)}, pages 464--472.

\bibitem[Song et~al., 2018]{song-etal-2018-leveraging}
Song, L., Wang, Z., Hamza, W., Zhang, Y., and Gildea, D. (2018).
\newblock Leveraging context information for natural question generation.
\newblock In {\em Proceedings of the 2018 Conference of the North {A}merican
  Chapter of the Association for Computational Linguistics: Human Language
  Technologies, Volume 2 (Short Papers)}, pages 569--574, New Orleans,
  Louisiana. Association for Computational Linguistics.

\bibitem[Su et~al., 2020]{su-etal-2020-multi}
Su, D., Xu, Y., Dai, W., Ji, Z., Yu, T., and Fung, P. (2020).
\newblock Multi-hop question generation with graph convolutional network.
\newblock In {\em Findings of the Association for Computational Linguistics:
  EMNLP 2020}, pages 4636--4647, Online. Association for Computational
  Linguistics.

\bibitem[Varga and Ha, 2010]{varga2010wlv}
Varga, A. and Ha, L.~A. (2010).
\newblock Wlv: a question generation system for the {QGSTEC} 2010 task b.
\newblock {\em Boyer \& Piwek (2010)}, pages 80--83.

\bibitem[Zhang and Bansal, 2019]{Zhang2019AddressingSD}
Zhang, S. and Bansal, M. (2019).
\newblock Addressing semantic drift in question generation for semi-supervised
  question answering.
\newblock In {\em EMNLP}.

\bibitem[Zhang et~al., 2020]{bert-score}
Zhang, T., Kishore, V., Wu, F., Weinberger, K.~Q., and Artzi, Y. (2020).
\newblock Bertscore: Evaluating text generation with bert.
\newblock In {\em International Conference on Learning Representations}.

\bibitem[Zhao et~al., 2018]{zhao-etal-2018-paragraph}
Zhao, Y., Ni, X., Ding, Y., and Ke, Q. (2018).
\newblock Paragraph-level neural question generation with maxout pointer and
  gated self-attention networks.
\newblock In {\em Proceedings of the 2018 Conference on Empirical Methods in
  Natural Language Processing}, pages 3901--3910, Brussels, Belgium.
  Association for Computational Linguistics.

\bibitem[Zhou et~al., 2018]{10.1007/978-3-319-73618-1_56}
Zhou, Q., Yang, N., Wei, F., Tan, C., Bao, H., and Zhou, M. (2018).
\newblock Neural question generation from text: a preliminary study.
\newblock In Huang, X., Jiang, J., Zhao, D., Feng, Y., and Hong, Y., editors,
  {\em Natural Language Processing and Chinese Computing}, pages 662--671,
  Cham. Springer International Publishing.

\bibitem[Zhou et~al., 2019]{Zhou2019QuestiontypeDQ}
Zhou, W., Zhang, M., and Wu, Y. (2019).
\newblock Question-type driven question generation.
\newblock In {\em EMNLP}.

\end{thebibliography}

\appendix

\end{document}